\documentclass{article} 
\PassOptionsToPackage{numbers, compress}{natbib}
\usepackage[preprint]{neurips_2024}

\usepackage{amsmath,amsfonts,bm}

\usepackage{amsthm}
\newcommand{\indep}{\perp\!\!\!\!\perp} 

\newtheorem{theorem}{Theorem}[section]
\newtheorem{lemma}{Lemma}[section]
\newtheorem{remark}{Remark}[section]
\newtheorem{definition}{Definition}[section]
\newtheorem{corollary}{Corollary}[section]
\newtheorem{proposition}{Proposition}[section]

\def\eqref#1{equation~\ref{#1}}

\def\1{\bm{1}}

\def\rmE{{\mathbf{E}}}

\def\rmS{{\mathbf{S}}}

\def\rmV{{\mathbf{V}}}

\def\rmX{{\mathbf{X}}}
\def\rmY{{\mathbf{Y}}}
\def\rmZ{{\mathbf{Z}}}

\DeclareMathAlphabet{\mathsfit}{\encodingdefault}{\sfdefault}{m}{sl}
\SetMathAlphabet{\mathsfit}{bold}{\encodingdefault}{\sfdefault}{bx}{n}

\DeclareMathOperator*{\argmax}{arg\,max}
\DeclareMathOperator*{\argmin}{arg\,min}

\usepackage{graphicx} 
\usepackage{wrapfig}
\usepackage{hyperref}
\usepackage{cleveref}
\usepackage{url}
\usepackage{xcolor}
\usepackage{bbm}
\usepackage{enumerate}
\usepackage{wrapfig}
\newcommand{\pref}[1]{(\ref{#1})}

\title{Robust Multi-view Co-expression Network Inference}

\author{Teodora Pandeva \\ University of Amsterdam \And Martijs Jonker \\ University of Amsterdam \And  Leendert Hamoen \\ University of Amsterdam \And Joris Mooij \\ University of Amsterdam \And Patrick Forr\'e \\ University of Amsterdam }

\begin{document}
\maketitle
\begin{abstract}
Unraveling the co-expression of genes across studies enhances the understanding of cellular processes. Inferring gene co-expression networks from transcriptome data presents many challenges, including spurious gene correlations, sample correlations, and batch effects. To address these complexities, we introduce a robust method for high-dimensional graph inference from multiple independent studies.  We base our approach on the premise that each dataset is essentially a noisy linear mixture of gene loadings that follow a multivariate $t$-distribution with a sparse precision matrix, which is shared across studies. This allows us to show that  we can identify the co-expression matrix up to a scaling factor among other model parameters. Our method employs an Expectation-Maximization procedure for parameter estimation. Empirical evaluation on synthetic and gene expression data demonstrates our method's improved ability to learn the underlying graph structure compared to baseline methods.  \end{abstract}

\section{Introduction}
\label{sec:intro}
\begin{wrapfigure}{r}{7cm}
    \centering
\includegraphics[width=\linewidth]{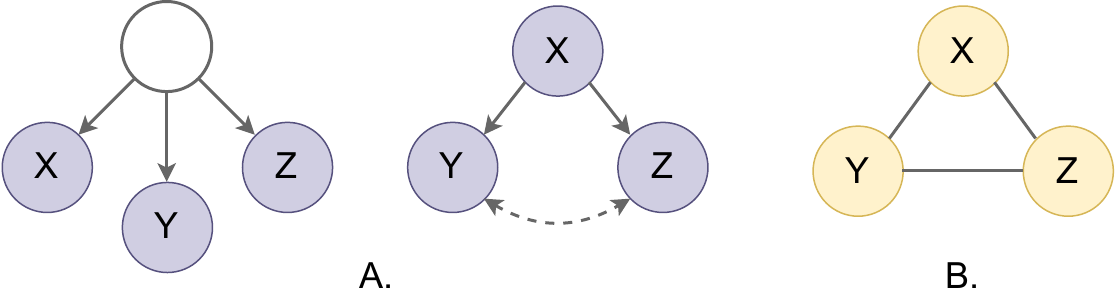}
    \caption{Two variations of the gene regulation of genes $X, Y, Z$ (A) colored in purple and their corresponding co-expression network illustrated in (B) in yellow. In (A) (left), genes $X,Y,Z$ are regulated by a common latent factor, such as another gene. The example in (A) (right) shows that gene $X$ regulates both $Y$ and $Z$. In addition, a bi-directional dashed line indicates potential confounding between genes $Y$ and $Z$.}
    \label{fig:coexp}
\end{wrapfigure}

  Over the past decades, advances in DNA sequencing technologies have led to significant advances in gene regulation research. These developments have provided deep insights into biological functions and disease processes. One notable example, which we will revisit later,  is the comprehensive study of the bacterium \textit{Bacillus subtilis}. This Gram-positive bacterium serves as a model organism for studying bacterial chromosome replication and cell differentiation. A substantial research endeavor has led to a continuous manual collection of biological findings about \textit{Bacillus subtilis} regulation and gene functionality on the online platform \textit{SubtiWiki} \citep{pedreira2021}, providing a clearer and more precise understanding of its cellular processes. This underscores the importance of developing methods that facilitate this process by robustly identifying such gene-gene interactions in a vast collection of experimental data from multiple sources, such as different technologies and laboratories.

Biologically relevant gene-gene interactions are often represented by a gene co-expression network (GCN), which is an undirected graph where each node corresponds to a gene. Genes that are  connected or positioned closely within the GCN  belong to the same functional modules, indicating that they work together to perform coordinated cellular activities. Therefore, constructing a GCN facilitates the understanding of gene regulation mechanisms. In this work, we aim to construct a GCN that closely resembles a gene regulatory network, considering only links that connect genes within the same regulatory network, such as regulator-regulated gene pairs or co-regulated genes (see \Cref{fig:coexp}).

Inferring GCNs from data can be very challenging, mainly due to hidden confounders due to the different experimental designs and batch effects associated with the different data sources. In response, current research in gene co-expression analysis often makes specific assumptions about the data generation model to deal with this complexity. This is typically represented by a noisy decomposition model: $\rmX=\rmS A +\rmE,$ where $\mathbf{X}\in\mathbb{R}^{p \times n}$ is a gene expression matrix describing the activity of $p$ genes across $n$ different samples (experiments, patients, tissues, etc.), $\rmS\in\mathbb{R}^{p \times k}$ is the \textit{gene loading matrix}, $A$  is the \textit{sample loading matrix}, and  $\rmE$ is the additive noise. A common assumption is that GCNs can be reconstructed from the gene loadings, where gene clusters are identified from each latent vector, a column in the gene loadings matrix $\rmS$ (e.g. \citep{moran2021spike,hochreiter2010fabia,sastry2019escherichia}).

This paper presents a novel probabilistic method for inferring complex network structures from high-dimensional data across multiple views. Unlike traditional approaches that rely primarily on clustering techniques or Gaussian models (e.g. \citep{moran2021spike,hochreiter2010fabia,gao2016context,kim2007sparse,danaher2014joint,guo2011joint}), our method employs a matrix-variate $t$-distribution framework that extends TLASSO by \citet{Finegold2011tlasso}. We refer to this extended model as MVTLASSO, which captures the covariance at both the sample and variable levels in the multi-view setting. Key contributions of this work, besides the proposed model, include the formulation of identifiability guarantees for the model parameters, such as the sparse precision matrix, which we can identify up to a scalar multiple (see \Cref{subsec:identifiability}). For model estimation, we implement an Expectation-Maximization (EM) procedure, which is described in \Cref{sec:em}. We apply MVTLASSO to both synthetic datasets and real-world gene expression data to validate its effectiveness. Our empirical results in \Cref{sec:experiments} show that MVTLASSO consistently demonstrates improved accuracy in reconstructing the underlying graph structures compared to baseline methods. %

\section{Robust Co-Expression Inference from non-i.i.d Samples}
In this section, we introduce and justify our chosen generative model, which we will refer to as \textbf{MVTLASSO}, placing it within the broader context of known GCN inference methods. In \Cref{subsec:identifiability}, we present theoretical guarantees for recovering the true model parameters. %

 Our approach can be seen as an instance of ICA, where the latent components, or gene loadings, are divided into two categories: those used to construct the GCN, denoted by $\rmS$, and those considered noise, denoted by $\rmZ$, which do not contribute to the GCN inference. We infer the GCN from the sparse precision matrix $\Theta$ estimated from \textit{all} ``useful'' gene loadings $\rmS$ across datasets (or \textit{views}) that follow a multivariate $t$-distribution similar to \citep{Finegold2011tlasso}.  More specifically, we make the following assumptions regarding the data generation process:
\begin{definition}
\label{def:model}
    Consider the scenario where we are given $D$ different data sets $\rmX_d\in \mathbb{R}^{p\times n_d}$, which may come from different sources and follow the representation:
\begin{align*}
    \rmX_d=\rmS_dA_d + \rmZ_dB_d
\end{align*}
where for each $d=1,\ldots, D$ it holds 
\begin{enumerate}
    \item   $(A_d^\top|B_d^\top)^\top \in \mathbb{R}^{(k_d+r_d)\times n_d}$  have full row rank with $\operatorname{rank}(A_d)=k_d$ and ${\operatorname{rank}(B_d)=n_d-k_d=:r_d,}$
    \item the columns of $\rmS_d\in\mathbb{R}^{p\times k_d}$ are mutually independent and follow a multivariate $t$-distribution, i.e. $\rmS_{d_{:,i}}\sim t_p(\nu, \mu_d, \Sigma)$ with $\nu>2$ degrees of freedom and  a sparse inverse dispersion matrix $\Theta:=(\Sigma)^{-1}$ that has a prior distribution  $p_\lambda(\Theta)$ with $\lambda>0$ defined as
\begin{align*}
    p_{\lambda}(\Theta)\propto \exp\left(-\lambda \|\Theta\|_1\right)\;\; \text{ with } \;\; \|\Theta\|_1=\sum_{i,j} |\Theta_{ij}|,
\end{align*}
\vspace{-0.5cm}
    \item the columns of $ \rmZ_d\in\mathbb{R}^{p\times r_d}$ are noise random variables and  are i.i.d multivariate $t$-distributed $t_p(\nu, 0, \sigma_d^2\mathbb{I}_p)$, such that there is no $\lambda \in \mathbb{R}$ with $ \sigma_d^2\mathbb{I}_p=\lambda \Sigma,$
    \item the latents  $\rmS_d$ and noise matrix $\rmZ_d$ are independent.
\end{enumerate}
\end{definition}

This perspective on $\Theta$ as a representation of the GCN closely aligns our work with that proposed by \citet{stegle2011efficient} for the single view case. Compared to \citep{stegle2011efficient}, we shift from a multivariate normal distribution to a multivariate $t$-distribution with sparse $\Theta$. Although this moves away from the theoretical guarantees of conditional independence to a more relaxed condition of conditional uncorrelation, as outlined by \citet{Finegold2011tlasso}, this approach provides more robust inference for unknown parameters, in this case, $A_d, B_d,\mu_d,\Theta$. This robustness is particularly beneficial in the presence of data contamination, a common challenge in the analysis of transcriptome data.

Finally, our model is reminiscent of dimensionality reduction methods, similar to the application of PCA, aiming to identify $k_d$ components per dataset (or view) that capture the most significant signals from the data. The remaining components are considered as i.i.d.\ noise, following a multivariate $t$-distribution, which does not play a role in estimating the network structure, represented by $\Theta$. A similar decomposition is proposed by \citet{parsana2019addressing}, where the authors show that removing the noise components after applying PCA improves the GCN inference of several algorithms.

\subsection{Identifiability Guarantees}
\label{subsec:identifiability}
Next, we will present our theoretical guarantees for identifying model parameters from \Cref{def:model}, i.e $\{A_d,B_d,\mu_d,\sigma^2_d\}$, $d=1,\dots,D$, and $\Sigma$. We will show that the  location $\mu_d$ and dispersion matrix $\Sigma$ of the gene loadings, as well as the sample loadings $A_d$ are identifiable up to the same constant across all views: 
\begin{proposition}
\label{prop:multi-view}
        Let $\rmX_1,\ldots, \rmX_D$ with $\rmX_d\in \mathbb{R}^{p\times n_d}$ be random matrices with the following two representations:
\begin{align*}
   \rmS_d^{(1)}A_d^{(1)} + \rmZ_d^{(1)}B_d^{(1)}= \rmX_d=\rmS_d^{(2)}A_d^{(2)} + \rmZ_d^{(2)}B_d^{(2)},
\end{align*}
where for $d=1,\ldots, D$, both representations $A_d^{(1)}\in\mathbb{R}^{k_d^{(1)}\times n_d},B_d^{(1)}\in\mathbb{R}^{ (n_d-k_d^{(1)})\times n_d},\rmS_d^{(1)}\in \mathbb{R}^{p\times k_d^{(1)}},\rmZ_d^{(1)}\in \mathbb{R}^{p\times (n_d-k_d^{(1)})}$
and
$A_d^{(2)}\in\mathbb{R}^{k_d^{(2)}\times n_d},B_d^{(2)}\in\mathbb{R}^{ (n_d-k_d^{(2)})\times n_d},\rmS_d^{(2)}\in \mathbb{R}^{p\times k_d^{(2)}}, \rmZ_d^{(2)}\in \mathbb{R}^{p\times (n_d-k_d^{(2)})}$  satisfy the properties  of \Cref{def:model}. Then, for $d=1,\ldots, D$, ${k_d^{(1)}=k_d^{(2)}=:k_d}$. Furthermore, there exist permutation matrices ${P_{A_1},\ldots, P_{A_D}, P_{B_1},\ldots, P_{B_D}}$ and constants $c, c_{1}, \ldots,  c_{D}> 0$ such that:
\begin{align*}
A_d^{(2)}&=cP_{A_d}A_d^{(1)},  &&  
 \Sigma_S^{(2)}=\frac{\Sigma_S^{(1)}}{c^2},&& \mu_{S_d}^{(2)}=\frac{\mu_{S_d}^{(1)}}{c}, \\
 B_d^{(2)}&=c_{d} P_{B_d}B_d^{(1)}, && \Sigma_{Z_d}^{(2)}=\frac{\Sigma_{Z_d}^{(1)}}{c_{d}^2},&& \mu_{Z_d}^{(2)}=\frac{\mu_{Z_d}^{(1)}}{c_{d}}.
\end{align*}
\end{proposition}

In contrast to well-established results in the ICA literature \citep{comon1994independent,kagan1973characterization}, which provide identifiability for the univariate case, we extend these results to multivariate elliptic distributions, as shown in \Cref{thm:multi-view}. \Cref{prop:multi-view} is a special case and a direct consequence of our more general results.

\section{Parameter Estimation}
\label{sec:em}

We begin by deriving the data likelihood, drawing inspiration from the ICA literature, e.g.\ the works of \citep{hyvarinen1999fast,amari1995new}. Instead of making derivations with respect to $A_d$ and $B_d$, we proceed in terms of the inverse of the concatenated matrix, denoted as ${W_d = (A_d \mid B_d)^{-1}}$. Consequently, the ``unmixed'' signal $\rmY_d := \rmX_dW_d$ represents the estimates for the latent vectors $\rmS_{d_{:,i}}$ for $i=1,\ldots, k_d$, and $\rmZ_{d_{:,i}}$ for $i=1,\ldots, n_d-k_d$, up to some scaling and permutation as described in \Cref{prop:multi-view}\footnote{Specifically, the first $k_d$ columns of $\rmY_d$ correspond to the estimates of $\rmS_{d}$, while the remaining $n-k_d$ columns correspond to the estimates of $\rmZ_{d}$.}. These signals follow a multivariate $t$-distribution. Thus, the likelihood for all views $\rmX_1,\ldots, \rmX_D$ is:
\begin{align}
\label{eq:likelihood}
    p(\rmX_1,\ldots, \rmX_D\mid \{\; W_d,\mu_d, \sigma_d \; \}_{d=1}^{D},\Sigma ) &= \prod_{d=1}^D p(\rmX_d) = \prod_{d=1}^D |\operatorname{det}W_d| \; p(\rmX_d\cdot W_d) \\\notag
    & = \prod_{d=1}^D |\operatorname{det}W_d |\prod_{i=1}^{n_d}t_{p}\left(\rmY_{d_{:,i}}\mid\nu,\rho_{d_i}, \Phi_{d_i}\right), 
\end{align}
where $\Phi_{d_i}=\mathbbm{1}_{\{i\leq k_d\}}\Sigma + \mathbbm{1}_{\{i> k_d\}} \sigma_d^2 \mathbb{I}_p$ and $\rho_{d_i}=\mathbbm{1}_{\{i\leq k_d\}}\mu_d$. Thus, the data likelihood is proportional to the product of the probabilities of $\sum_{d=1}^D n_d$ independent multivariate $t$-distributed vectors.

\subsection{The Expectation-Maximization Procedure}
Unfortunately, directly estimating the unknown parameters from \pref{eq:likelihood} is infeasible. However, we can leverage the alternative representation of the multivariate $t$-distribution described in \Cref{thm:gen-t}, which is central to the EM procedure proposed by \citet{liu1995ml,Finegold2011tlasso}. For each random vector $\rmY_{d_{:,i}}$, the corresponding generative process can equivalently be represented as:
\begin{align*}
    \tau_{d_i}&\sim \Gamma \left(\frac{\nu}{2}, \frac{\nu}{2}\right), &&
    \rmY_{d_{:,i}}\sim \mathcal{N}(\rho_{d_i}, \Phi_{d_i}/\tau_{d_i}),
\end{align*}
where the variables $\tau_{d_i}$ are unobserved. Thus, the complete data log-likelihood  with unknown parameters ${\bm{\gamma}:=\{ W_d, \mu_d, \sigma_d\}_{d=1}^D\cup \{\Sigma\}}$ and random variables $\rmX_1,\ldots, \rmX_D$ and $\tau_1,\ldots \tau_D$ with ${\tau_d:=(\tau_{d_1}, \ldots, \tau_{d_{n_d}})}$ is given by 
\begin{align}
\label{eq:loglikelihood}
    l\left(\bm{\gamma};\;\{\rmX_d,\tau_d\}_{d=1}^D\right)
    \propto \sum_d &\Bigg\{\ln|\operatorname{det}W_d |+\sum_{i=1}^{n_d}\frac{1}{2}\ln\det \Phi_{d_i}^{-1} -\frac{\tau_{d_i}}{2}\operatorname{tr}\left(\Phi_{d_i}^{-1}\rmY_{d_{:,i}}\rmY_{d_{:,i}}^\top\right)\\ \notag
    &+\tau_{d_i}\rho_{d_i}^\top\Phi_{d_i}^{-1}\rmY_{d_{:,i}}-\frac{\tau_{d_i}}{2}\rho_{d_i}^\top\Phi_{d_i}^{-1}\rho_{d_i}\Bigg\},
\end{align}
where $\Phi_{d_i}^{-1} = \mathbbm{1}_{\{i\leq k_d\}}\Theta + \mathbbm{1}_{\{i> k_d\}} \frac{1}{\sigma_d^2} \mathbb{I}_p$ with $\Theta=(\Sigma)^{-1}$. The right side of \pref{eq:loglikelihood} is linear in the latent variables $\tau_{d_i}$. Thus, for the E-step it suffices to compute $\mathbb{E}[\tau_{d_i}\mid \rmX_d]$ for every $d=1,\ldots,D$ and $i=1,\ldots, n_d.$ This can be derived directly by observing that the conditional distribution ${p( \tau_{d_i}\mid \rmX_d) = p( \tau_{d_i}\mid \rmY_{d_{:,i}})}$ is given by
\begin{align*}
    \tau_{d_i}\mid \rmY_{d_{:,i}}\sim\Gamma\left(\frac{\nu+p}{2}, \dfrac{\nu+\delta(\rmY_{d_{:,i}}, \rho_{d_i},\Phi_{d_i})}{2}\right)
    \end{align*}
with
    \begin{align*}
    \delta(\rmY_{d_{:,i}}, \rho_{d_i},\Phi_{d_i}) &= (\rmY_{d_{:,i}}-\rho_{d_i})^\top \Phi_{d_i}^{-1}(\rmY_{d_{:,i}}-\rho_{d_i}).
\end{align*}
Consequently, for the conditional expectation we get: ${\mathbb{E}[\tau_{d_i}\mid \rmY_{d_{:,i}}]=\dfrac{\nu+p}{\nu+\delta(\rmY_{d_{:,i}}, \rho_{d_i},\Phi_{d_i})}}$. 

Hence, the EM procedure iterates through two main steps for each view $d$: 1) the estimation of $\tau_{d_i}$ while keeping $\rho_{d_i}$, $\Phi_{d_i}$, and $W_d$ fixed; and 2) the estimation of $\rho_{d_i}$, $\Phi_{d_i}$, $W_d$, and $\Theta := (\Sigma)^{-1}$, where $\Theta$ is determined by solving the graphical lasso (GLASSO) problem as described by \citep{friedman2008sparse}. This method is designed to estimate sparse precision matrices in a multi-view setting. 
The EM procedure at step $t \geq 1$ is performed as follows:

\paragraph{E-step:} For fixed estimated  $\mu_d^{(t-1)},\Sigma^{(t-1)},\sigma_d^{(t-1)}$and $W_d^{(t-1)}$ compute $\mathbb{E}[\tau_{d_i}\mid \rmX_d]$, i.e.
\begin{align*}
    \rmY_d^{(t-1)} &= \rmX_dW_d^{(t-1)}, &&
    \tau_{d_{i}}^{(t)} = \dfrac{\nu+p}{\nu+\delta(\rmY_{d_{:,i}}^{(t-1)}, \rho_{d_i}^{(t-1)},\Phi_{d_i}^{(t-1)})}.
\end{align*}
\paragraph{M-step:} Solve the optimization problem:
\begin{align*}
   \bm{\gamma}^{(t)} \in\argmax_{\bm{\gamma}}l\left(\bm{\gamma};\;\{\rmX_d,\tau_d^{(t)}\}_{d=1}^D\right),
\end{align*}
with $\bm{\gamma}^{(t)} = \{ W_d^{(t)}, \mu_d^{(t)}, \sigma_d^{(t)}\}_{d=1}^D\cup\{\Sigma^{(t)}\}$ that leads to the following steps for all $d=1,\ldots, D$:
\begin{enumerate}
    \item Calculate  $\mu_d^{(t)},\Sigma^{(t)}$ and $\sigma_d^{(t)}$ for fixed $\tau_{d_{i}}^{(t)}$ and $\rmY_d^{(t)}$
\begin{align*}
    \mu_d^{(t)} &= \dfrac{\sum_{i=1}^{k_d} \tau_{d_i}^{(t)}\rmY_{d_{:,i}}^{(t-1)}}{\sum_{i=1}^{k_d} \tau_{d_i}^{(t)}}, 
   \;\;\;\; \Sigma^{(t)} = \frac{1}{\sum_d k_d} \sum_d\sum_{i=1}^{k_d} \tau_{d_i}^{(t)} \left(\rmY_{d_{:,i}}^{(t-1)}-\mu_d^{(t)}\right)\left(\rmY_{d_{:,i}}^{(t-1)}-\mu_d^{(t)}\right)^\top, \\
     \sigma_d^{(t)}& = \sqrt{\frac{1}{p(n-k_d)} \sum_{i=k_d+1}^{n} \sum_{l=1}^{p}\tau_{d_i}^{(t)} (\rmY_{d_{l,i}}^{(t-1)})^2}
\end{align*}
\item Estimate $\Theta$ via solving the GLASSO optimization problem for $\Sigma^{(t)}$ with penalty parameter $\lambda>0$ given by:
\begin{align}
\label{eq:opt-glasso}
   \Theta^{(t)}\in \argmin_{\Theta\succ 0 } -\ln\det(\Theta) + \operatorname{tr}(\Sigma^{(t)}\Theta)+\lambda \Vert \Theta\Vert_1
\end{align}
\item Estimate $W_d^{(t)}$ for fixed $\mu_d^{(t)},\Sigma^{(t)}, \sigma_d^{(t)}$ and $\tau_{d_{i}}^{(t)}$:
\begin{align}
\label{eq:opt-W}
    W_d^{(t)} \in \argmin_{W} \Big\{ &\operatorname{tr}\left(\left(\rmX_dW - \bm{\mu}^{(t)}_d\right)^\top\Theta^{(t)}\left(\rmX_dW - \bm{\mu}^{(t)}_d\right)\mathcal{T}_1^{(t)}\right) \\ \notag
    &+ \frac{1}{(\sigma_d^{(t)})^2}\operatorname{tr}\left(W^\top \rmX_d^\top\rmX_d W\mathcal{T}_2^{(t)}\right) -\ln|\operatorname{det}W| \Big\},
\end{align}
where $\bm{\mu}^{(t)}_d:=(\underbrace{{\mu}^{(t)}_d,\ldots,{\mu}^{(t)}_d}_{k_d},0\ldots, 0)\in \mathbb{R}^{p\times n_d},$ and  $\mathcal{T}_1^{(t)}, \mathcal{T}_2^{(t)}\in \mathbb{R}^{n_d\times n_d}$ are diagonal matrices defined as  $\mathcal{T}_1^{(t)}=\operatorname{diag}(\tau_{d_{1}}^{(t)},\ldots, \tau_{d_{k_d}}^{(t)},0,\ldots, 0)$ and  ${\mathcal{T}_2^{(t)}=\operatorname{diag}(0,\ldots, 0,\tau_{d_{k_d+1}}^{(t)},\ldots, \tau_{d_{n_d}}^{(t)})}$
\end{enumerate}

\begin{figure}
    \centering
    \includegraphics[width=\columnwidth]{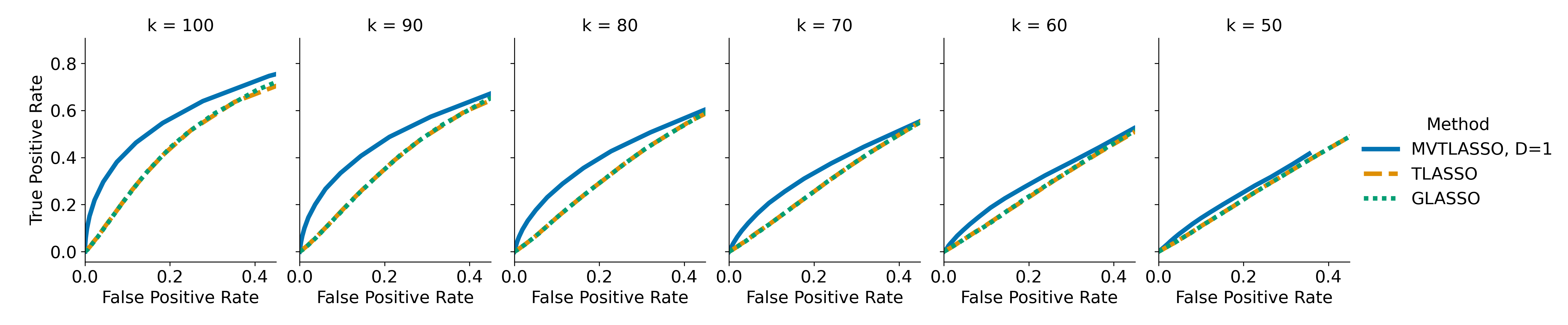}
    \caption{ROC curves summarizing the benchmark experiment on the data generated as described in \Cref{subsec:sim} with a total of $100$ sample loadings. Each curve represents the average result of $100$ experiments. The number of signal loadings $k$ varies in each experiment, as indicated in the subplot titles. The results show that MVTLASSO outperforms TLASSO and GLASSO.
}
    \label{fig:sim-benchmark}
\end{figure}
Details on the implementation of the EM procedure can be found in \Cref{app:implementation}.

\begin{wrapfigure}{r}{0.4\textwidth}
    \centering
    \includegraphics[width=0.3\textwidth]{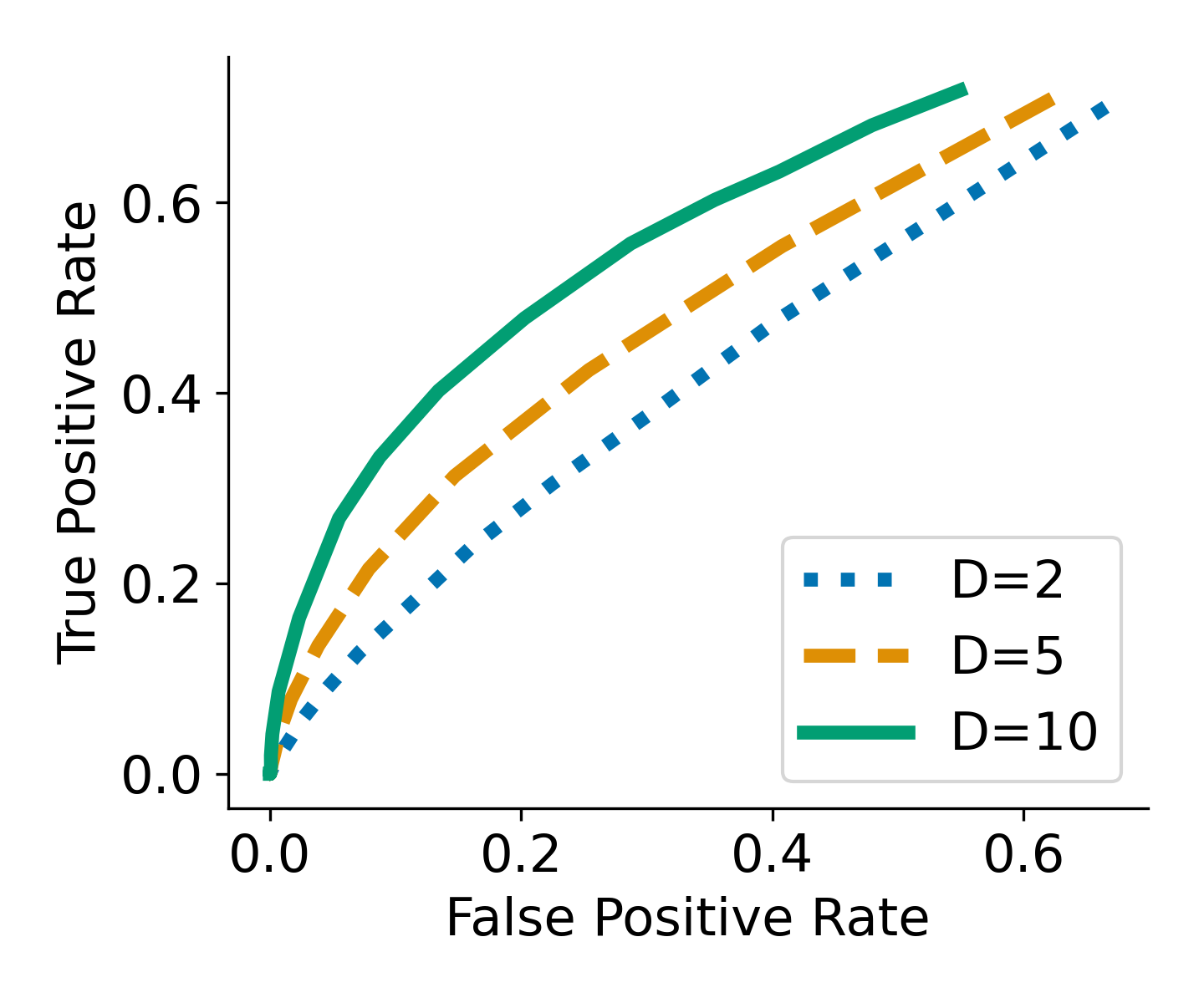}
    \caption{We set both $k=50$ and $r=50$ and varied $D=2, 5, 10$. For each case, we ran 100 synthetic experiments while varying the sparsity parameter. The averaged ROC curves show that increasing the number of views ($D$) improves performance.}
    \label{fig:sim-multiview}
    \vspace{-12pt}
\end{wrapfigure}

\section{Results}
\label{sec:experiments}

\subsection{Simulated Data}
\label{subsec:sim}

We benchmark our method against \textbf{GLASSO} \citep{friedman2008sparse} and \textbf{TLASSO} \citep{Finegold2011tlasso} using simulated data whose generative model aligns with \Cref{def:model} and follows a similar setup to that proposed by  \citet{Finegold2011tlasso}, with $200$ variables and $100$ samples.  The sparse precision matrix $\Theta$ is generated as follows 1) off-diagonal entries $\Theta_{ij}$ with $i\neq j$ are sampled from $\{-1,0,1\}$ with probabilities $\{0.01,0.98, 0.01\}$ 2) the diagonal entries are set to 1 plus the number of edges connected with the node, i.e.  $\Theta_{ii}=1+\sum_{j}\mathbbm{1}_{\{\Theta_{ij}\neq 0\}}$. Additionally, we set $\mu=0$ and $\sigma=1$ in all experiments. The sample loading matrices $A$ and $B$ have entries sampled according to standard normal distribution.

In the single view case ($D=1$), we evaluated our model and several baselines on datasets with varying ratios of $k$ signal loadings ($\rmS$) to $r$ noise loadings ($\rmZ$): ${100:0, 90:10, 80:20, 70:30, 60:40,}$ and $50:50.$  We ran 100 experiments for each method for 50 sparsity parameters $\lambda$. For each experiment, we calculated the true positive and false positive rates individually and then averaged these  for each sparsity parameter.  

The aggregated results are shown in \Cref{fig:sim-benchmark}. Across all scenarios indicated by the subplot titles, MVTLASSO, as expected, consistently outperforms the baselines. However, it is evident that the performance of all methods decreases as the proportion of noise loadings increases. \Cref{fig:sim-multiview}  illustrates that with an increasing number of views $D$, where the ratio of signal loadings $k$ to noise loadings $r$ is $50:50$, the performance of MVTLASSO improves significantly. Specifically, the aggregated true positive rate from 100 experiments increases relative to the false positive rate.

\subsection{Gene Co-Expression Inference for Bacillus Subtilis}
\begin{wrapfigure}{r}{0.4\textwidth}
    \centering
    \includegraphics[width=0.4\textwidth]{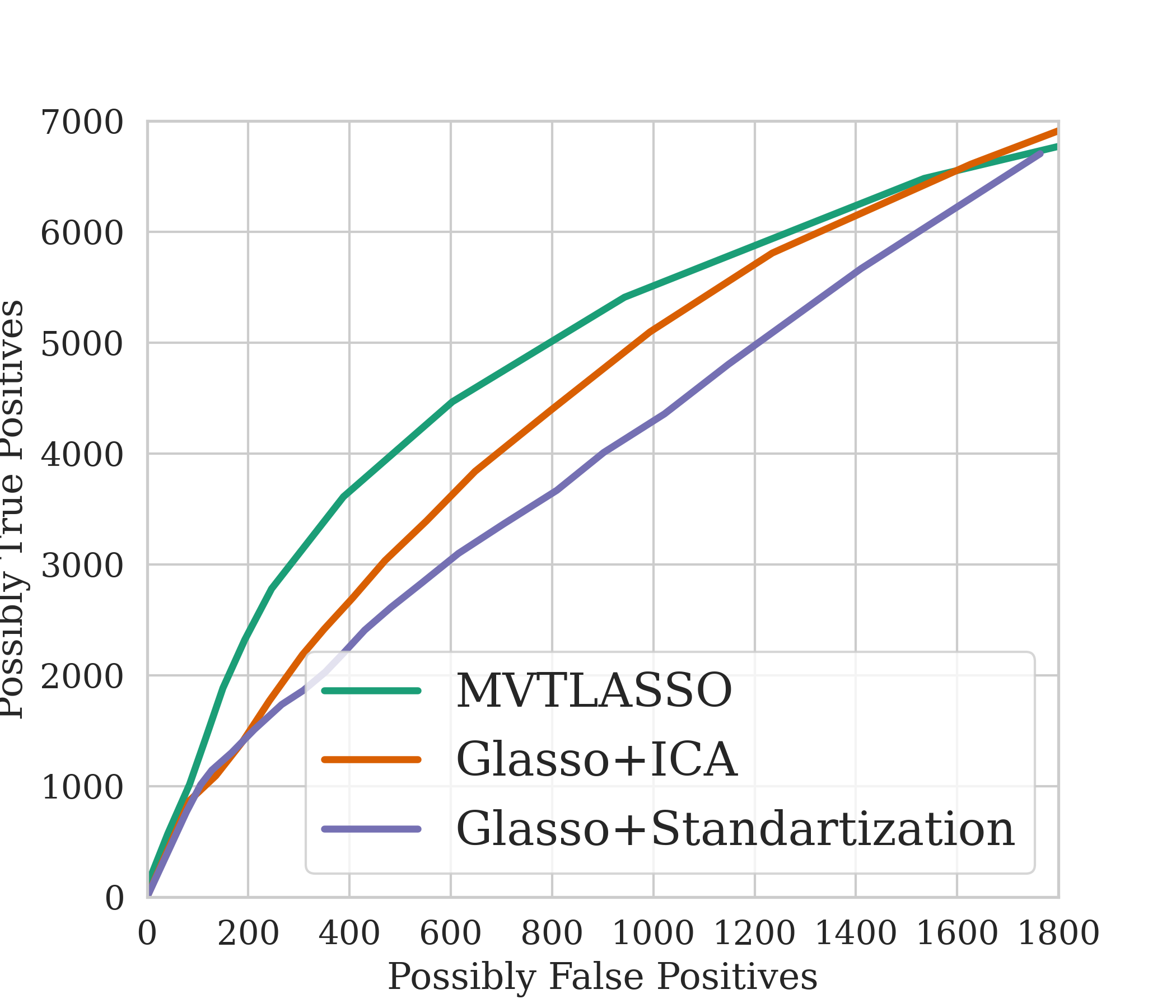}
    \caption{True positive vs. possibly false positive edges obtained via stability selection for various penalty parameters. The results demonstrate that MVTLASSO consistently infers more true positive edges across all settings.}
    \label{fig:b-subtilis}
\end{wrapfigure}
\vspace{-7.4pt}
We revisit the motivational example of the GCN inference from  \textit{B. Subtilis} gene expression data. For this purpose,  we use two well-controlled transcriptome data compendia. These datasets were collected using the \textit{B. subtilis} strain BSB1, which contains 269 samples from 104 different experimental conditions \citep{nicolas2012condition}, and the closely related strain PY79, which contains data from 38 unique experimental designs \citep{arrieta2015experimentally}. The \textit{B. Subtilis} genome contains approximately 4100 genes, and for every transcriptome experiment, gene expressions from   3994 genes were obtained.  Both datasets include a wide range of conditions, including growth in different media, competence, biofilm formation, swarming, different stress conditions, sporulation, and knockout experiments. The data were preprocessed as outlined in  \Cref{app:datapreprop}. We further split each dataset into two approximately equal subsets of samples, ensuring they are as distinct as possible in their experimental design.

We then used these four views to benchmark MVTLASSO against two other methods: GLASSO+ICA and GLASSO+Standardization, described in \Cref{app:implementation}. Unfortunately, the correct number of gene loadings $k_d$ remains unknown. We estimated $k_d$ following the approach in  \citep{parsana2019addressing} prior to fitting the models. 
The fitting process for all methods includes stability selection as described by \citet{meinshausen2010stability} and detailed in \Cref{app:implementation}. In this approach, for each of 15 penalty parameters,``stable'' edges are selected from 100 precision matrices, each estimated from bootstrapped samples containing $90\%$ of all data. For each penalty parameter, we count the true positive edges, as verified against the ground truth data from  \textit{SubtiWiki}, as well as potential false positives from all selected edges. The true positive vs false positive counts for each method are shown in  \Cref{fig:b-subtilis}. These results indicate that MVTLASSO consistently identifies more true positive edges across most penalty parameters compared to the other two methods.

\section{Discussion}
We introduced MVTLASSO, a robust method for inferring gene co-expression networks from high-dimensional gene expression data across multiple independent studies. Our approach effectively addresses the inherent complexity of gene expression data, including gene and sample correlations as well as batch effects, by modeling each dataset as a noisy linear mixture of gene loadings governed by a multivariate t-distribution with a sparse precision matrix. We employ an EM procedure for parameter estimation, supported by theoretical guarantees that ensure the identifiability of the model parameters. Empirical evaluations on both synthetic and real gene expression data have demonstrated the superior performance of MVTLASSO compared to baseline methods. Our method consistently shows improved accuracy in learning the underlying graph structures, underscoring its robustness and reliability.

A promising direction for future work is to develop a more efficient and reliable hyperparameter selection procedure.  The selection of sample dimensions and noise loadings can be challenging and time-consuming due to the implemented EM procedure. In addition, incorporating available experimental metadata into the modeling process could provide further refinement and improve the overall performance of MVTLASSO.

\newpage
\bibliography{main}
\bibliographystyle{abbrvnat}
\newpage
\appendix
\section{Related Work} Inferring GCNs from data can be very challenging, mainly due to hidden confounders and batch effects associated with the different data sources. In response, current research in gene co-expression analysis often makes specific assumptions about the data generation model to deal with this complexity. This is typically represented by a noisy decomposition model: $\rmX=\rmS A +\rmE,$ where $\mathbf{X}\in\mathbb{R}^{p \times n}$ is a gene expression matrix describing the activity of $p$ genes across $n$ different samples (experiments, patients, tissues, etc.), $\rmS\in\mathbb{R}^{p \times k}$ is the \textit{gene loading matrix}, $A$  is the \textit{sample loading matrix}, and  $\rmE$ is the additive noise. These approaches can be broadly categorized into decomposition methods and their refinements, biclustering algorithms. 

\textit{Decomposition methods}, including Independent Component Analysis (ICA), Principal Component Analysis (PCA), and other variations of factor analysis, have shown remarkable effectiveness in identifying clusters of genes connected in the GCN. These methods are used to analyze single data sets \citep{saelens2018comprehensive,rychel2020machine} as well as to integrate data from multiple studies \citep{le2008sparse,smilde2017common,kim2017,pandeva2023multi}. A common assumption is that GCNs can be reconstructed from the gene loadings, where gene clusters are identified from each latent vector, a column in the gene loadings matrix $\rmS$, usually by thresholding. Often, these clusters are assumed to represent sets of genes connected within the GCN and mapped to gene modules with a common function.

\textit{Biclustering algorithms} aim to cluster genes and samples simultaneously by applying sparsity constraints to both gene and sample loadings, e.g., \citep{moran2021spike,hochreiter2010fabia,gao2016context,kim2007sparse}, providing a principled approach for a two-fold clustering. This approach assumes that the sample loading matrix $A$ will have a sparse pattern, i.e., only a small group of genes will deviate within a small subset of samples.  These methods are particularly useful for subgroup analyses, such as classifying patients into different subtypes based on gene expression levels. 

Despite their ability to cluster, all these methods do not model the relationships between clusters and thus do not provide a comprehensive strategy for inferring gene co-expression graphs. One exception is the Kronecker graphical LASSO approach by \citep{stegle2011efficient}, which constructs a sparse graph structure while modeling sample covariance. However, this method has not been extended to handle multiple datasets collected from different labs and may lack robustness against data contamination. On the other hand, existing methods that use the graphical LASSO to infer GCNs from various data sources \citep{danaher2014joint,guo2011joint}  do not address the confounding variables in the experiments and assume that the data are independent and identically distributed.
\section{Identifiability}
In our analysis, we will make use of the multivariate elliptical distributions, denoted by  $E_p(\mu,\Sigma)$,  whose density $f(x;\mu, \Sigma)$ is proportional to $f(x;\mu, \Sigma)\propto g((x-\mu)^\top \Sigma (x-\mu))$ for some measurable function $g$ and a positive semi-definite dispersion matrix $\Sigma$ and median $\mu$. An example of such elliptical distributions is the Gaussian and multivariate $t$-distribution. First, we show that the sample loadings are identifiable up to scaling and permutation, provided that none of the gene loadings have Gaussian marginals. This result is an extension of Theorem 10.3 in \citep{kagan1973characterization} for the multivariate case:
\begin{lemma}
\label{lemma:gaussian}
  Let $\mathbf{X}\in \mathbb{R}^{p\times n}$ be a random matrix. Assume the following two representations of $\mathbf{X}$
    \begin{align*}
       \mathbf{S}^{(1)} A^{(1)}=\mathbf{X} = \mathbf{S}^{(2)}A^{(2)},
    \end{align*}
     with the following properties for $i=1,2:$
     \begin{enumerate}
         \item $A^{(i)}\in \mathbb{R}^{k^{(i)}\times n}$ is a (non-random)  matrix with a full row rank
         \item  $\mathbf{S}^{(i)}\in \mathbb{R}^{p\times k^{(i)}}$ is a random matrix such that the columns of $\mathbf{S}^{(i)}$ are mutually independent.
     \end{enumerate}
     If the $i$-th row of $A^{(1)}$ is not proportional to any row of $A^{(2)}$ then the $i$-th column of $\rmS^{(1)}$ has Gaussian distributed marginals. Additionally, if the $i$-th column of $\rmS^{(1)}$ follows an elliptical distribution, then it is multivariate Gaussian.
\end{lemma}

\paragraph{Proof of \Cref{lemma:gaussian}}
{\begin{proof}
    W.l.o.g. let $i=1$. According to \citep[Lemma 10.2.2]{kagan1973characterization}   there exists a $n\times 2$ matrix $H$ such that the matrices $C_1 =A^{(1)}H$ and $C_2 = A^{(2)}H$ of orders $k^{(1)}\times 2$ and $k^{(2)}\times 2$ respectively have the following property; the first row of $C_1$ is not proportional to any of the other rows of $C_1$ or to any of the rows of $C_2.$ 
    
    Now consider the following algebraic relationship for $\mathbf{Y} =\mathbf{X}H$:
    \begin{align*}
         \mathbf{S}^{(1)} C_1=\mathbf{Y}  = \mathbf{S}^{(2)}C_2,
    \end{align*}
    where $\mathbf{Y}\in \mathbb{R}^{p\times 2}$. For each row $r=1,\ldots p$ of  $\mathbf{Y}$ we have the two equivalent representations
    \begin{align*}
        \mathbf{S}^{(1)}_{r,:} C_1=\mathbf{Y}_{r,:}  = \mathbf{S}^{(2)}_{r,:}C_2.
    \end{align*}
    Thus, by \citep[Lemma 10.2.4]{kagan1973characterization}, it follows that $\mathbf{S}^{(1)}_{r,1}$ is Gaussian distributed because the first row of $C_1$ is not proportional to any of the other rows of $C_1$ or to the one of $C_2.$ Consequently, this implies that the marginal distributions of $\mathbf{S}^{(1)}_{:,1}$ are Gaussians. Given that $\mathbf{S}^{(1)}_{:,1}$ is elliptical with the previous argument it follows that it is Gaussian \citep{fang2018symmetric}.
\end{proof}
\begin{theorem}
\label{thm:ident-of-noiseless-case}
    Let $\mathbf{X}\in \mathbb{R}^{p\times n}$ be a random matrix. Assume the following two representations of $\mathbf{X}$
    \begin{align*}
       \mathbf{S}^{(1)} A^{(1)}=\mathbf{X} = \mathbf{S}^{(2)}A^{(2)}
    \end{align*}
    with the following properties for $i=1,2:$
    \begin{enumerate}
        \item $A^{(i)}\in \mathbb{R}^{k^{(i)}\times n}$ is a (non-random) matrix with full row rank, i.e. $\operatorname{rank}(A^{(i)})=k^{(i)}$
        \item  $\mathbf{S}^{(i)}\in \mathbb{R}^{p\times k^{(i)}}$ is a random matrix such that
        \begin{enumerate}
            \item The columns of $\mathbf{S}^{(i)}$ are mutually independent, 
                        \item For  $k=1,\ldots,k^{(i)}$ the vectors $\mathbf{S}^{(i)}_{:,k}$ are distributed according to a non-Gaussian elliptical distribution $ E_p(\mu^{(i)},\Sigma^{(i)})$ with mean $\mu^{(i)}$ and a dispersion matrix  $\Sigma^{(i)}$.
            \item Additionally, $\mathbf{S}^{(i)}_{:,k}$ the random vectors do not have Gaussian components.
        \end{enumerate}
    \end{enumerate}
            Then $k^{(1)}=k^{(2)}=k$ and there exist a 
        permutation matrix $P=P(\rho)\in \mathbb{R}^{k\times k}$ given by $Pe_j=e_{\rho(j)}$ 
        and a constant $c> 0$ such that:
        \begin{align*}
            A^{(2)}&=c PA^{(1)}, && \Sigma^{(2)}=\frac{\Sigma^{(1)}}{c^2}&& \mu^{(2)}=\frac{\mu^{(1)}}{c}.
        \end{align*}
\end{theorem}}
{\begin{proof}
    \Cref{lemma:gaussian} establishes that each row of matrix $A^{(1)}$ is proportional to a row of $A^{(2)}$. Now if we assume that $k^{(1)}>k^{(2)}$ then there must be at least two distinct rows in  $A^{(1)}$ that are proportional to the same row of $A^{(2)}$. This contradicts the assumption that both $A^{(1)}$ and $A^{(2)}$ have full row rank. Thus, it follows that $k^{(1)}=k^{(2)}=:k$ and  there exist a permutation matrix $P\in \mathbb{R}^{k\times k}$ and an invertible diagonal matrix $\Lambda=\operatorname{diag}(\lambda_1, \ldots, \lambda_k)\in \mathbb{R}^{k\times k}$ such that $ A^{(2)}=\Lambda PA^{(1)}$. 

Note that for the characteristic function of a matrix $\mathbf{S}$ that fulfills 2a) to c) for some mean $\mu$ and dispersion matrix $\Sigma$ holds
    \begin{align*}
        \chi_{\mathbf{S}}(\mathbf{t}) &= \mathbb{E}\left[\exp(i \operatorname{tr}(\mathbf{t}^\top \mathbf{S}))\right]=  \mathbb{E}\left[\exp\left(i \sum_j \mathbf{t}_{:,j}^\top \mathbf{S}_{:,j})\right)\right] \\
        &= \prod_j \chi_{\mathbf{S}_{:,j}}\left(\mathbf{t}_{:,j}\right)=\prod_j\chi_{\mathbf{S}_{:,1}}\left(\mathbf{t}_{:,j}\right)\\
        &=\prod_j \exp\left(i\mathbf{t}_{:,j}^\top \mu\right)\psi\left(\mathbf{t}_{:,j}^\top\Sigma \mathbf{t}_{:,j}\right),
    \end{align*}
where $\psi$ is the characteristic generator and   $\mathbf{t}\in \mathbb{R}^{p\times k}$. 

Let $\Tilde{\mathbf{S}}^{(2)} = \mathbf{S}^{(2)}P$. Then we get for the characteristic functions of $\Tilde{\mathbf{S}}^{(2)}$ and ${\mathbf{S}}^{(1)}$ for all $\mathbf{t}\in \mathbb{R}^{p\times k}$  $\chi_{\mathbf{S}^{(1)}}(\mathbf{t})=\chi_{\Tilde{\mathbf{S}}^{(2)}\Lambda}(\mathbf{t})$
\begin{align*}
    \prod_j &\exp\left(i\mathbf{t}_{:,j}^\top \mu^{(1)}\right)\psi_1\left(\mathbf{t}_{:,j}^\top\Sigma^{(1)} \mathbf{t}_{:,j}\right)\\
    &=  \prod_j \exp\left(i \lambda_j\mathbf{t}_{:,j}^\top \mu^{(2)}\right)\psi_2\left(\lambda_j^2\mathbf{t}_{:,j}^\top\Sigma^{(2)} \mathbf{t}_{:,j}\right),
\end{align*}
where $\psi_i$ is the characteristic generator corresponding to the $i-$the representation. Consequently,  for each $j$ with $\mathbf{t}_{:,j}=t\in \mathbb{R}^{p}$ and otherwise $\mathbf{t}_{:,r}=0$ for all $r\neq j$ we get
\begin{align*}
    \exp&\left(it^\top \mu^{(1)}\right)\psi_1\left(t^\top\Sigma^{(1)} t\right)\\
    &= \exp\left(i \lambda_jt^\top \mu^{(2)}\right)\psi_2\left(\lambda_j^2t^\top\Sigma^{(2)} t\right)
\end{align*}
It follows that $\lambda_1=\ldots=\lambda_k=c$ and $\mu^{(1)}=c\mu^{(2)}$, and $\Sigma^{(1)}=c^2\Sigma^{(2)}.$
\end{proof}}
Next, we show that by imposing additional constraints on the gene loadings - in particular, requiring that they come from the same elliptic non-Gaussian multivariate distribution - it becomes possible to determine that the sample matrix, along with its locations and dispersion matrix, are identifiable up to a scalar:
\begin{theorem}
\label{thm:single-view}
        Let $\mathbf{X}\in \mathbb{R}^{p\times n}$ be a random matrix. Assume the following two representations of $\mathbf{X}$
    \begin{align*}
    \mathbf{S}^{(1)} A^{(1)}+ \mathbf{Z}^{(1)} B^{(1)}=\mathbf{X} = \mathbf{S}^{(2)}A^{(2)}+ \mathbf{Z}^{(2)} B^{(2)}
    \end{align*}
    with the following properties for $i=1,2:$
    \begin{enumerate}
        \item $(A^{(i)\top}|B^{(i)\top})^\top\in \mathbb{R}^{(k^{(i)}+l^{(i)})\times n}$ is a (non-random) matrix with full row rank with $\operatorname{rank}(A^{(i)})=k^{(i)}$ and $\operatorname{rank}(B^{(i)})=l^{(i)}\leq n-k^{(i)}$
        \item  $\mathbf{S}^{(i)}\in \mathbb{R}^{p\times k^{(i)}}$  and $\mathbf{Z}^{(i)}\in \mathbb{R}^{p\times l^{(i)}}$ are random matrices such that for $i=1,2$ and $\rmV^{(i)}\in\{\mathbf{S}^{(i)}, \mathbf{Z}^{(i)}\}$
        \begin{enumerate}
            \item The columns of $\rmV^{(i)}$ are mutually independent, 
            \item The column vectors $\rmV^{(i)}_{:,k}$ are distributed according to a non-Gaussian elliptical distribution $ E_p(\mu_{V}^{(i)},\Sigma_{V}^{(i)})$ with location $\mu_{V}^{(i)}$ and a dispersion matrix  $\Sigma_{V}^{(i)}$.
            \item Additionally, the random column vectors of $\rmS_d$ and $\rmZ_d$ do not have Gaussian components.
        \end{enumerate}
        \item the latents  $\rmS^{(i)}$ and noise matrix $\rmZ^{(i)}$ are independent and there exist no $\lambda\in\mathbb{R}$ such that $\mu_Z^{(i)}=\lambda\mu_S^{(i)}, \Sigma_Z^{(i)}=\lambda^2\Sigma_S^{(i)} $.
    \end{enumerate}
            Then $k^{(1)}=k^{(2)}=k$ and $l^{(1)}=l^{(2)}=l$ and exist 
        permutation matrices $P_A, P_B\in \mathbb{R}^{k\times k}$
        and constants $c_A, c_B> 0$ such that:
        \begin{align*}
            A^{(2)}&=c_A P_AA^{(1)},  && B^{(2)}=c_B P_BB^{(1)}, \\
            \Sigma_S^{(2)}&=\frac{\Sigma_S^{(1)}}{c_A^2},&& \mu_S^{(2)}=\frac{\mu_S^{(1)}}{c_A},\\
            \Sigma_Z^{(2)}&=\frac{\Sigma_Z^{(1)}}{c_B^2},&& \mu_Z^{(2)}=\frac{\mu_Z^{(1)}}{c_B}.
        \end{align*}
\end{theorem}
{\paragraph{Proof of \Cref{thm:single-view}}
\begin{proof} 
    According to \Cref{lemma:gaussian} each row of $(A^{(1)\top}|B^{(1)\top})^\top$ is proportional to a row of $(A^{(2)\top}|B^{(2)\top})^\top$. With similar arguments as above it holds that $k^{(1)}+l^{(1)}=k^{(2)}+l^{(2)}$. 
    
    Suppose that the $j-$th row of $A^{(1)}$ is proportional to the $r-$th row of $B^{(1)}.$ It follows that that there exist a constant $\lambda$ such that for all $t\in\mathbb{R}^p:$
    \begin{align*}
    \exp\left(it^\top \mu_S^{(1)}\right)\psi_1\left(t^\top\Sigma_S^{(1)} t\right)= \exp\left(i \lambda t^\top \mu_Z^{(2)}\right)\psi_2\left(\lambda ^2t^\top\Sigma_Z^{(2)} t\right),
\end{align*}
    i.e., $\mu_S^{(1)} = \lambda\mu_Z^{(2)}$ and $\Sigma_S^{(1)} = \lambda^2\Sigma_Z^{(2)}.$
    Thus, $k^{(1)}=k^{(2)}$ and  $l^{(1)}=l^{(2)}$. The rest follows from \Cref{thm:ident-of-noiseless-case}.
\end{proof}}
\begin{corollary}
\label{thm:multi-view}
        Let $\rmX_1,\ldots, \rmX_D$ with $\rmX_d\in \mathbb{R}^{p\times n_d}$ be random matrices with the following two representations
\begin{align*}
   \rmS_d^{(1)}A_d^{(1)} + \rmZ_d^{(1)}B_d^{(1)}= \rmX_d=\rmS_d^{(2)}A_d^{(2)} + \rmZ_d^{(2)}B_d^{(2)}
\end{align*}
with the following properties for $i=1,2$ and $d=1,\ldots, D:$
\begin{enumerate}
    \item $(A_d^{(i)\top}|B_d^{(i)\top})^\top\in \mathbb{R}^{(k_d^{(i)}+l_d^{(i)})\times n_d}$ is a (non-random) matrix with full row rank:
    \begin{align*}\operatorname{rank}(A_d^{(i)})&=k_d^{(i)},&& \operatorname{rank}(B_d^{(i)})=l_d^{(i)}\leq n_d-k_d^{(i)},
    \end{align*}
    \item the columns of $\rmS_d^{(i)}$ are independent and are distributed according to a non-Gaussian elliptical distribution $ E_p(\mu_{S_d}^{(i)},\Sigma_{S_d}^{(i)})$ with location $\mu_{S_d}^{(i)}$ and a dispersion matrix  ${\Sigma_{S}^{(i)}:=\Sigma_{S_1}^{(i)}=\ldots=\Sigma_{S_D}^{(i)}}$.
    \item the columns of $ \rmZ_d^{(i)}$ are noise random variables and  are i.i.d non-Gaussian elliptical distributed $ E_p(\mu_{Z_d}^{(i)},\Sigma_{Z_d}^{(i)})$ with location $\mu_{Z_d}^{(i)}$ and a dispersion matrix  $\Sigma_{Z_d}^{(i)}$. Furthermore, for each $d$ there exist no $\lambda\in\mathbb{R}$ such that $\mu_{Z_d}^{(i)}=\lambda\mu_S^{(i)}, \Sigma_{Z_d}^{(i)}=\lambda^2\Sigma_S^{(i)} $.
    \item the latents  $\rmS_d^{(i)}$ and noise matrix $\rmZ_d^{(i)}$ are mutually independent.
     \item Additionally, the random column vectors of $\rmS_d$ and $\rmZ_d$  do not have Gaussian components.
\end{enumerate}

Then, for $d=1,\ldots, D$, $k_d^{(1)}=k_d^{(2)}=k_d$ and $l_d^{(1)}=l_d^{(2)}=l_d$. Furthermore, there exist permutation matrices ${P_{A_1},\ldots, P_{A_D}, P_{B_1},\ldots, P_{B_D}}$ and constants $c_A, c_{B_1}, \ldots,  c_{B_D}> 0$ such that:
\begin{align*}
A_d^{(2)}&=c_A P_{A_d}A_d^{(1)},  && B_d^{(2)}=c_{B_d} P_{B_d}B_d^{(1)}, \\
\Sigma_S^{(2)}&=\frac{\Sigma_S^{(1)}}{c_A^2},&& \mu_{S_d}^{(2)}=\frac{\mu_{S_d}^{(1)}}{c_A}, \\
\Sigma_{Z_d}^{(2)}&=\frac{\Sigma_{Z_d}^{(1)}}{c_{B_d}^2},&& \mu_{Z_d}^{(2)}=\frac{\mu_{Z_d}^{(1)}}{c_{B_d}}.
\end{align*}

\end{corollary}
\paragraph{Proof of \Cref{thm:multi-view}}
\Cref{thm:single-view} guarantees the identifiability results for each view separately, i.e. for each $d=1,\ldots, D$ there exist permutation matrices ${ P_{A_d}, P_{B_d}}$ and constants $c_{A_d}, c_{B_d}> 0$ such that:
\begin{align*}
A_d^{(2)}&=c_{A_d} P_{A_d}A_d^{(1)},  && B_d^{(2)}=c_{B_d} P_{B_d}B_d^{(1)}, \\
\Sigma_S^{(2)}&=\frac{\Sigma_S^{(1)}}{c_{A_d}^2},&& \mu_{S_d}^{(2)}=\frac{\mu_{S_d}^{(1)}}{c_{A_d}}, \\
\Sigma_{Z_d}^{(2)}&=\frac{\Sigma_{Z_d}^{(1)}}{c_{B_d}^2},&& \mu_{Z_d}^{(2)}=\frac{\mu_{Z_d}^{(1)}}{c_{B_d}}.
\end{align*}
It follows that for all $d=1,\ldots, D:$
\[\Sigma_S^{(2)}=\frac{\Sigma_S^{(1)}}{c_{A_d}^2}.\]
Thus, $c_A:=c_{A_1}=\ldots=c_{A_D}$.
\section{Dependence Structure and Properties of the Multivariate t-Distribution}
\subsection{Alternative Generative Model for the Multivariate t-Distribution}
The probability density function of the multivariate \( t \)-distribution with \( \nu \) degrees of freedom, mean vector \( \mathbf{\mu} \), and scale matrix \( \mathbf{\Sigma} \) in \( p \) dimensions is given by:

\[
f(\mathbf{x}) = \frac{\Gamma\left(\frac{\nu + p}{2}\right)}{\Gamma\left(\frac{\nu}{2}\right) (\nu \pi)^{p/2} |\mathbf{\Sigma}|^{1/2}} \left(1 + \frac{1}{\nu} (\mathbf{x} - \mathbf{\mu})^T \mathbf{\Sigma}^{-1} (\mathbf{x} - \mathbf{\mu})\right)^{-\frac{\nu + p}{2}}
\]

where:
\begin{itemize}
    \item \( \mathbf{x} \) is the variable vector,
    \item \( \mathbf{\mu} \) is the mean vector,
    \item \( \mathbf{\Sigma} \) is the scale matrix,
    \item \( \nu \) is the degrees of freedom,
    \item \( \Gamma \) is the gamma function.
\end{itemize}

The following result is central to the EM procedure and is shows that the multivariate $t$-distribution can be expressed by means of the multivariate normal distributed random variable and Gamma distributed random variable:
\begin{theorem}[\citep{arellano1995some}]
\label{thm:gen-t}
    Let $S \sim t_p(\nu,\mu,\Sigma)$ for some mean $\mu$ and positive semi-definite matrix $\Sigma$. Then, there exist random variables $\tau$ and $N$ that follow  Gamma distribution $\Gamma(\frac{\nu}{2},\frac{\nu}{2})$ and a Gaussian distribution $\mathcal{N}(0,\Sigma),$ respectively, such that $S\sim \mu + N/\sqrt{\tau}.$
\end{theorem}

\subsection{Dependence Relationship between the Genes in the GCN}
\label{subsec:dependence}

As previously discussed, we use the precision matrix $\Theta$ to construct the GCN. Specifically,  consider a graph $G = (V, E)$, where $V = \{1, \ldots , p\}$ represents the set of observed genes and $E$ is a collection of  edges between pair of nodes (or genes) $i$ and $j$ for which the corresponding entry  $\Theta_{{ij}}$ is non-zero.

An interesting aspect is understanding the types of (in)dependencies encoded by this graph structure. For context, in Gaussian models, the absence of an edge between two nodes $i$ and $j$ implies conditional independence between them, given the remaining nodes. However, this direct implication does not translate to multivariate $t$-distributions. Instead, a weaker concept of dependence, conditional uncorrelation, applies, as discussed  in \citep{Finegold2011tlasso} for the single view case:
\begin{theorem}[\citep{Finegold2011tlasso}] 
\label{thm:uncorr}
Let $S\sim t_{p}(\nu,\mu,\Sigma).$  $\Sigma$ is a positive definite matrix with $(\Sigma^{-1})_{ij}=0$  for indices $i\neq j$ corresponding to the non-edges in the graph $G$. If two nodes $i$ and $j$ are separated by a set of nodes $C$ in $G$, then $S_i$ and $S_j$ are conditionally uncorrelated given $S_C.$
\end{theorem}
While \Cref{thm:uncorr} shows that conditional uncorrelation can be derived from the graph structure, it leaves open the question of whether multivariate $t$-distributions can be factorized according to any Bayesian network. The following result addresses this issue by showing that the only Bayesian network compatible with the multivariate $t$-distribution is a fully connected DAG:
\begin{lemma}
\label{lemma:imp}
    Let $G=(V,E)$ be a DAG with vertices $V=\{1, \ldots, p\}$. Furthermore, the joint distribution of the corresponding variables $S_1,\ldots, S_p$ is multivariate $t$-distribution $ t_{p}(\nu,\mu,\Sigma)$ with $0<\nu<\infty.$ Let $\operatorname{pa}(k)\subseteq V\setminus\{k\}$ denote the set of parents of node $k$. Then, the following holds $P(S_1,\ldots, S_p) = \prod_{k=1}^p P(S_k\mid S_{\operatorname{pa}(k)}) $ iff there exists an ordering $S_{\tau(1)},\ldots, S_{\tau(p)}$ such that ${\operatorname{pa}(\tau(k))=\{\tau(1),\ldots, \tau(k-1)\}}$, i.e. the graph is fully connected.
    
\end{lemma}
\begin{remark}
    \begin{enumerate}[(a)]
        \item \Cref{lemma:imp} suggests that from the estimated $\Theta$, we can infer only conditional uncorrelation between the genes, not conditional independence. However, this result does not contradict the GCN definition used in this work, as detailed in \Cref{sec:intro}, which is based on correlation rather than statistical independence. 
        \item   According to \Cref{thm:uncorr}, the reconstructed GCN should exclude edges between genes that are conditionally uncorrelated  given rest of the genes.  This implies that co-regulated genes will not be connected in the GCN, as they become conditionally uncorrelated when conditioned on their regulators. However, this holds only in the absence of confounding factors—whether observed or unobserved—such as external stimuli that might influence gene regulation as part of the experimental design. Our current approach does not account for these elements of the experimental design, which could potentially refine the GCN. We leave this consideration for future work.
    \end{enumerate}
\end{remark}

\paragraph{Proof of \Cref{lemma:imp}}
{\begin{proof}
    $``\Leftarrow"$ This direction follows directly from the chain rule of probabilities. 

     $``\Rightarrow"$ Assume that the DAG is not fully connected, i.e. there exist  sets $A, B, C\subset V,\; A \neq \emptyset, B\neq \emptyset$ such that the random variables $S_A$  and $S_B$ are d-separated given $S_C$ (${S_A\perp_{G}S_B|S_C}$). Thus, it follows that ${S_A\indep S_B \mid S_C}$ which implies that $p(S_A\mid S_B, S_C)=p(S_A|S_C).$ 
     
     According to \citep{arellano1995some} the joint distribution of $S_A,S_B, S_C$, their conditionals and marginals follow a multivariate $t$-distribution. More precisely, let $d=|A|+|B|+|C|,\;\mu_d = (\mu_A^\top, \mu_B^\top,\mu_C^\top)^\top,\; \Sigma = \Sigma_{(A,B,C),(A,B,C)}$, then ${S_d=(S_A,S_B, S_C)\sim t_d(\nu,\mu_d,\Sigma)}$. Furthermore, for the conditional distributions we have
     \begin{align}
     \label{eq:cond-t}
         S_A\mid S_B, S_C&\sim t_{|A|}\left(\nu+|B|+|C|,\mu_{A|B,C},\frac{\nu+d_{B,C}}{\nu+|B|+|C|}\Sigma_{A|B,C}\right)\\
         S_A\mid S_C&\sim t_{|A|}\left(\nu+|C|,\mu_{A|C},\frac{\nu+d_{C}}{\nu+|C|}\Sigma_{A|C}\right) \notag
     \end{align}
     Then, it follows that $\nu+|B|+|C|=\nu+|B|+|C|$ which implies that $|B|=0.$
\end{proof}}

\section{Parameter Inference: Background}

\subsection{Graphical LASSO}
\label{sec:glasso}
The Graphical lasso (GLASSO) is a maximum likelihood estimator for inferring graph structure within high-dimensional multivariate normal distributed data through estimating a sparse precision matrix  \citep{friedman2008sparse}.   More precisely, let $\mathbf{X}=(\mathbf{X}_1, \mathbf{X}_2, \ldots, \mathbf{X}_n)$ be a collection of $n$ i.i.d. samples distributed according to the multivariate normal distribution $\mathcal{N}(0, \Theta^{-1})$, where ${\Theta^{-1}\in\mathbb{R}^{p\times p}}$ is the covariance matrix and its inverse $\Theta$ known as  the precision matrix. The underlying undirected graph structure among the variables can be inferred directly from the precision matrix: a non-zero entry $\Theta_{ij}$ indicates an undirected edge between the $i$-th and $j$-the variables in the multivariate vector. GLASSO estimates $\Theta$ by maximizing the posterior distribution of $\mathbf{X}$ given $\Theta:=\Sigma^{-1}$ which is proportional to
\begin{align*}
    p(\mathbf{X}, \Theta) = p_\lambda(\Theta)\prod_{i=1}^n \mathcal{N}(\mathbf{X}_i|\mu, \Theta^{-1})\;\; \text{ where } \;\; \Theta\succ 0. 
\end{align*}
The prior $p_\lambda(\Theta)$ on the positive-definite matrices $\Theta$ parametrized by $\lambda>0$ is defined as
\begin{align*}
    p_{\lambda}(\Theta)\propto \exp\left(-\lambda \|\Theta\|_1\right)\;\; \text{ with } \;\; \|\Theta\|_1=\sum_{i,j} |\Theta_{ij}|.
\end{align*}
 Thus, the MLE problem that GLASSO solves can be formalized as follows
\begin{align}
\label{eq:glasso}
\max_{\Theta\succ 0 }\ln p(\mathbf{X}, \Theta) \equiv \min_{\Theta\succ 0 } -\ln\det(\Theta) + \operatorname{tr}(\hat\Sigma\Theta)+\lambda \Vert \Theta\Vert_1,
\end{align}
where $S$ is the empirical covariance matrix, $\hat\Sigma=\frac{1}{n}\sum_{i=1}^n (\mathbf{X}_i-\bar{\mathbf{X}})(\mathbf{X}_i-\bar{\mathbf{X}})^\top$, and $\bar{\mathbf{X}}$ is the empirical mean. Intuitively, the parameter $\lambda$ controls the sparsity level of the precision matrix $\Theta$. Specifically, selecting a higher value for $\lambda$ leads to sparser precision matrix estimates.  
\subsection{Student's t-Lasso}
The accuracy of graph inference can be significantly compromised by deviations from the normal distribution assumption.  To address this robustness issue, \citep{Finegold2011tlasso} propose an alternative to GLASSO for inferring graph structure of multivariate Student's $t$-distribution which we call TGLASSO. Consider the setting from above, where we are given a collection of $n$ i.i.d samples $\mathbf{X}=(\mathbf{X}_1, \mathbf{X}_2, \ldots, \mathbf{X}_n)$. Then the joint distribution of the data $\mathbf{X}$ and precision matrix $\Theta$ is given by
\begin{align*}
    p(\mathbf{X}, \Theta) = p_\lambda(\Theta)\prod_{i=1}^n t_{\nu, p}(\mathbf{X}_i|\mu, \Theta^{-1})\;\; \text{ where } \;\; \Theta\succ 0,
\end{align*}
where the density function of the Student's $t$-distribution $t_{\nu, p}(\mu, \Theta^{-1})$ is given by
\begin{align*}
    \frac{\Gamma\left((\nu+p)/2\right)\det \Theta^{1/2}}{(\pi \nu)^{p/2}\Gamma(\nu/2)\left(1+\delta\left(\mathbf{x};\mu, \Theta\right)/\nu\right)^{(\nu+p)/2}} 
\end{align*}    
with
   \begin{align*} 
    \delta\left(\mathbf{x};\mu, \Theta\right) = (\mathbf{x}-\mu)^\top \Theta (\mathbf{x}-\mu),\;\;\mathbf{x}\in \mathbb{R}^p.
\end{align*}
Estimating the precision matrix in this setting is not tractable, and \citep{Finegold2011tlasso} propose an Expectation-Maximization procedure for estimating $\Theta$ by exploiting the following generative model with latent variables $\mathbf{Z}_i$ and $\tau_i$ for each sample $\mathbf{X}_i$
\begin{align*}
    \mathbf{Z}_i&\sim \mathcal{N}(0, \Theta^{-1})\\
    \tau_i & \sim \Gamma(\nu/2, \nu/2)\\
    \mathbf{X}_i:=\mu + \mathbf{Z}_i/\sqrt{\tau}_i&\sim t_{\nu, p}(\mu, \Theta^{-1}).
\end{align*}
The proposed EM procedure operates under the assumption that  $\tau_i$'s are latent variables and that $\mathbf{X}_i|\tau_i\sim \mathcal{N}(\mu, (\tau_i\Theta)^{-1})$. This process iterates through two main steps:  1) Estimating the $\tau_i$ for fixed $\mu$ and $\Theta^{-1}$ and 2) Estimating $\mu$ and $\Theta^{-1}$, where $\Theta$ is a solution to the GLASSO problem in \Cref{eq:glasso} for an empirical covariance matrix of the estimated $\mathbf{Z}$. More precisely, at step $t\geq 0$ the EM procedure becomes

\paragraph{E-step:} For fixed estimated $\mu^{(t-1)}$ and $\Theta^{(t-1)}$ compute
\begin{align*}
    \tau_i^{(t)} = \dfrac{\nu+p}{\nu+\delta\left(\mathbf{X}_i;\mu^{(t-1)}, \Theta^{(t-1)}\right)}
\end{align*}
\paragraph{M-step:} Calculate $\mu^{(t)}$ and $\Sigma^{(t)}$
\begin{align}
    \mu^{(t)} &= \dfrac{\sum_{i=1}^n \tau_i^{(t)}\mathbf{X}_i}{\sum_{i=1}^n \tau_i^{(t)}} && \Sigma^{(t)} = \frac{1}{n} \sum_{i=1}^n \tau_i^{(t)} \left(\mathbf{X}_i-\mu^{(t)}\right)\left(\mathbf{X}_i-\mu^{(t)}\right)^\top
\end{align}
Estimate $\Theta^{(t)}$ via solving the GLASSO optimization problem
\begin{align*}
   \Theta^{(t)}\in \argmin_{\Theta\succ 0 } -\ln\det(\Theta) + \operatorname{tr}(\Sigma^{(t)}\Theta)+\lambda \Vert \Theta\Vert_1
\end{align*}
\section{Experiments}
\subsection{Implementation}
\label{app:implementation}
\paragraph{Implementation of MVTLASSO}
\begin{enumerate}
    
    \item    The optimization problem in step 3 is convex  when $\{W_d\}_{d=1}^D$ are positive semi-definite matrices. In the general case, we only require $\{W_d\}_{d=1}^D$ to be invertible which makes  solving \eqref{eq:opt-W} more challenging.  However, by treating the datasets $\rmX_1, \ldots, \rmX_D$ as instances of ICA  we can simplify the original problem \eqref{eq:opt-W}. By applying FastICA \citep{hyvarinen1999fast} or another ICA algorithm, we can obtain $\tilde\rmX_1, \ldots, \tilde\rmX_D$, which are estimates of $\rmY_1, \ldots, \rmY_D$, up to permutation matrices $P_1, \ldots, P_D$ and scaling given by diagonal matrices $\Lambda_1, \ldots, \Lambda_D$. Consequently, using $\tilde\rmX_1, \ldots, \tilde\rmX_D$ instead of the raw data transforms the original optimization problem \eqref{eq:opt-W} for each $W_d$ to finding $\tilde W_d = P_d\Lambda_d$. In our implementation, we jointly optimize for $\Lambda_d$ and $P_d$, with a further relaxation for $P_d$ to be orthogonal to speed up computations.
    \item We initialize the parameter $W_0$ as detailed in the previous remark (a), and the parameters $\mu_d, \sigma_d$ and $\Theta$  using a few iterations (not necessarily to convergence) of TLASSO by \citet{Finegold2011tlasso}.
    \item    The steps in the M-step are interdependent, i.e., steps 1 and 2 are based on the inverse sample matrix $W_d$. Therefore, it is possible to iterate steps 1 to 3 multiple times. In our implementation, however, we perform only a single iteration.
    \item Our model relies on several hyperparameters, including the number of multivariate $t$-distributed vectors $k_d$ used for graph inference and the penalty parameters $\lambda$ and $\gamma$. Ideally, these parameters could be determined by cross-validation. However, our EM procedure involves a GLASSO step in each iteration, which is computationally intensive. Therefore, we preselect the number of components prior to the parameter estimation process as described by \citet{parsana2019addressing}. 
\end{enumerate}
\paragraph{Implementation of baselines}
For GLASSO, we used the implementation available in the R package \citep{glassorpackage}. The variants GLASSO+Standardization and GLASSO+ICA include preliminary steps where the samples are subjected to standardization and ICA, respectively, before GLASSO is applied for precision matrix estimation. It is important to note that neither baseline includes a dimensionality reduction step.
\paragraph{Stability Selection}
The fitting procedure for all GLASSO-based methods makes use of stability selection by \citet{meinshausen2010stability} with a predefined range of penalty parameters. The steps of the procedure are outlined as follows:
\begin{enumerate}
\item  The data is repeatedly subsampled by selecting $90\%$ of all samples per view $N=100$ times. For each subsample, the selected GCN inference method is applied using the predefined set of penalty parameters, $\Lambda$.

\item The outcomes for each penalty parameter are gathered in the selection probability matrix $\Pi_\lambda$, where $(\Pi_\lambda)_{ij}$ represents the proportion of the $N$ precision matrices $\hat{\Theta}^{(1)}, \ldots, \hat{\Theta}^{(N)}$ indicating a nonzero edge between nodes $i$ and $j$, i.e.  $(\Pi_\lambda)_{ij}=\frac{\sum_{l}\mathbbm{1}_{\{\hat{\Theta}^{(l)}_{ij}\neq 0\}}}{N}$.

\item  We select the edges whose selection probability exceeds $50\%$ for each penalty parameter. 

\item The final graph can be constructed by collecting all edges inferred from the range of penalty parameters. This step is not conducted in our analysis. 
 \end{enumerate}
The primary benefit of stability selection, as outlined by \citep{meinshausen2010stability}, is that it can reduce the risk of false positives, i.e., incorrectly identifying edges in the network. By requiring that an edge be consistently identified across many subsamples of the data, stability selection ensures that the edges selected are robust and not the result of random variations in the data. 

\subsection{Data Preprocessing}
\label{app:datapreprop}
The dataset BSB1 is preprocessed following the method suggested by \citet{rychel2020machine}. Specifically, three samples (S3\_3, G+S\_1, and Mt0\_2) were removed to ensure that the Pearson correlation between biological replicates was at least 0.9.  Furthermore, we centered the data by subtracting the mean gene values in the M9 exponential growth condition. We used the preprocessed PY79 dataset by \citet{arrieta2015experimentally}.  BSB1 and PY79 samples are then centered and rescaled before applying any graph inference procedures. We selected genes that are present in both datasets. In addition, we have split both datasets into two subsets of samples with experimental designs that are as different as possible to simulate four views instead of two. A link to the datasets will be provided after submission.

\end{document}